\pdfoutput=1
\documentclass[11pt]{article}
\usepackage{booktabs}
\usepackage{tabularx}
\usepackage{array}
\usepackage{multirow}
\usepackage{makecell}
\usepackage{amsmath}
\usepackage{csquotes}
\usepackage{enumitem}
\usepackage{acl}
\usepackage{multirow}
\usepackage{times}
\usepackage{latexsym}
\usepackage{booktabs}
\usepackage[T1]{fontenc}
\usepackage{tabularx}
\usepackage[utf8]{inputenc}

\usepackage{microtype}

\usepackage{inconsolata}

\usepackage{graphicx}

\title{EmoS: A High-Fidelity Multimodal Benchmark for Fine-grained Streaming Emotional Understanding}

\author{
Pengze Guo\textsuperscript{1}\thanks{Equal contribution.}\quad
Jingxi Liang\textsuperscript{1}\footnotemark[1]\quad
Zhiwen Xie\textsuperscript{1,2}\footnotemark[2]\quad
Qifeng Wang\textsuperscript{1}\quad
Derek F. Wong\textsuperscript{1}\thanks{Corresponding author.}\\
\textsuperscript{1}$\spadesuit$ NLP\textsuperscript{2}CT Lab, Department of Computer and Information Science, University of Macau\\
\textsuperscript{2}School of Computer Science, Central China Normal University\\
\texttt{nlp2ct.\{pengze,jingxi,qifeng\}@gmail.com}, \texttt{zwxie@ccnu.edu.cn} \\
\texttt{derekfw@um.edu.mo}
}

\begin{document}
\maketitle
\begin{abstract}
In the context of today's high-pressure, aging society, the demand for large-scale emotional models capable of providing empathetic support is more critical than ever. However, existing benchmarks fail to simultaneously achieve ecological validity, signal clarity, and reliable fine-grained labeling. We introduce EmoS, a high-fidelity bilingual benchmark designed to resolve the limitations of ecological validity and noise in existing datasets by combining strictly filtered static slices with a dynamic Streaming Monologue subset. Supported by a rigorous dual-layer human annotation pipeline, EmoS provides trusted ground truth that captures continuous emotional evolution. Empirical results show that fine-tuning MLLMs (multimodal large language models) on EmoS yields significant gains over zero-shot baselines, laying the foundation for the training and evaluation of future emotion recognition models and empathy models. The dataset and code are publicly available at \url{https://github.com/NLP2CT/EmoS}.
\end{abstract}

\section{Introduction}

With the rapid advancement of artificial intelligence (AI) , there is an increasing societal expectation for AI agents to transcend simple task execution and act as an emotionally intelligent partner~\citep{cheng2024emotionllama,voila2025,liu2022chsims2}. In high-pressure environments, such as psychological counseling and aging care, AI systems must not only recognize static emotions but also proactively intervene before a user's state deteriorates. However, the development of such empathetic models is currently bottlenecked by the quality of underlying data. We identify a critical ``data quality trilemma'' in Multimodal Emotion Recognition (MER), where existing benchmarks struggle to simultaneously achieve ecological validity, signal clarity, and reliable labeling. Prior datasets often fail to meet these evolving needs. Early lab-controlled datasets like IEMOCAP~\citep{busso2008iemocap} and DAIC-WOZ \citep{gratch2014daic}, while clean, lack the spontaneity of real-life interactions. Conversely, Widely used 'in-the-wild' datasets like MELD~\citep{poria2019meld} and CH-SIMS v2 \citep{liu2022chsims2} provide rich contexts but are fundamentally compromised by unreliable and coarse-grained ground truth. Beyond simple modality noise, MELD suffers from heavy context dependence and ambiguity, while CH-SIMS v2 is limited to sentiment polarity rather than specific emotions, rendering both insufficient for fine-grained emotion recognition tasks. More critically, most datasets fragment continuous dialogues into isolated utterance-level snippets. This fragmentation severs the emotional evolution timeline, making it impossible for models to learn dynamic trajectories, such as the gradual escalation from frustration to rage. While recent synthetic datasets generated by Large Language Models (LLMs) \citep{cheng2024emotionllama,lian2025affectgpt} attempt to scale up data, they often suffer from hallucinated labels, compromising the credibility of the ground truth.

\begin{figure}[t]
    \centering
    \includegraphics[width=1\linewidth]{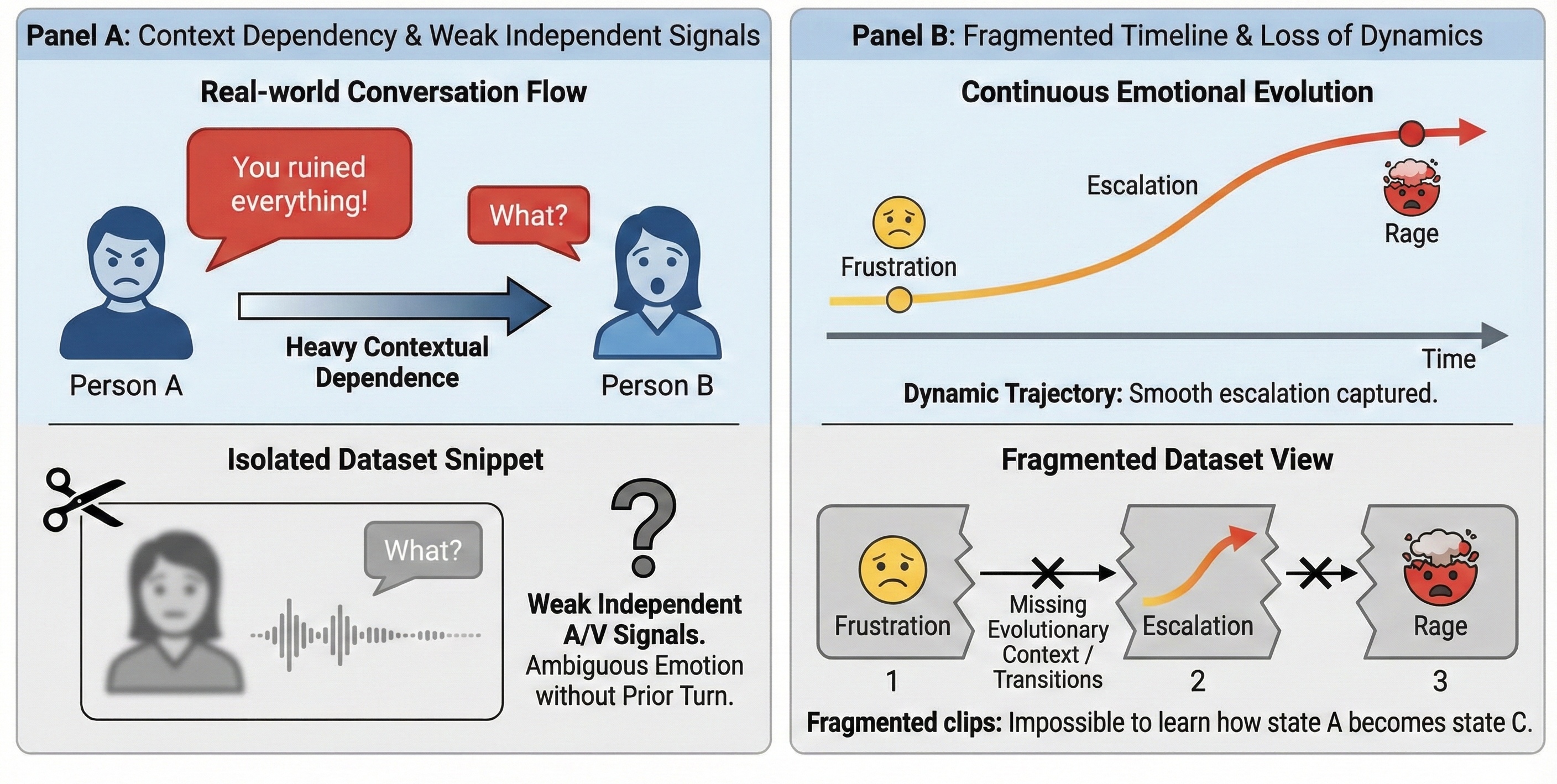}
    \caption{Illustration of limitations in current short-utterance multimodal datasets.}
    \label{fig:placeholder}
\end{figure}

The limitations of existing work underscore the critical necessity of constructing a high-quality MER benchmark. An ideal MER benchmark requires high-fidelity multimodal alignment across textual, acoustic, and visual signals. 
Crucially, it must provide fine-grained emotion labels (e.g., joy, anger) rather than coarse sentiment polarity. 
Furthermore, it should support continuous temporal modeling over long-form dialogues, moving beyond isolated utterance-level classification. To achieve this, we introduce \textbf{EmoS}, a high-fidelity bilingual benchmark spanning 9,403 static samples and 2 hours of streaming monologues. Constructed through a rigorous dual-layer annotation pipeline (Basic-7~\citep{ekman1992argument} and GoEmotions-28~\citep{demszky-etal-2020-goemotions}) validated by the Dawid-Skene algorithm~\citep{dawid1979maximum}, EmoS integrates a strictly filtered MELD-Core, fine-grained CH-SIMS v2, and a novel streaming subset to capture continuous emotional evolution.

We benchmarked SOTA MLLMs (e.g., Gemini-3, Qwen-3) on EmoS. Results indicate that Zero-Shot performance is limited, with Gemini-3 achieving only $\sim$61\% accuracy due to conservative neutral bias. Conversely, Task-Adaptive Fine-Tuning is essential: fine-tuning Qwen-3 boosted accuracy to 70.3\%, dramatically improving recall on long-tail emotions like Disgust (F1 0.30$\rightarrow$0.75). Furthermore, models with ultra-long context windows demonstrated exceptional sensitivity to narrative flow, successfully predicting 82\% of emotion turning points in the streaming subset.
In summary, our contributions are threefold:

\begin{itemize}[leftmargin=*, nosep, topsep=2pt]
    \item We present EmoS, a strictly cleaned, human-annotated benchmark ($N=9,403$ static samples + 2h streaming) that resolves modality noise and introduces a novel streaming subset.
    \item We establish a high-standard Dual-Layer Annotation Protocol combining basic and fine-grained taxonomies, supported by rigorous annotator style analysis.
    \item We provide a comprehensive Benchmark of SOTA MLLMs, showing that fine-tuning on high-quality data is prerequisite for mastering dynamic emotional trajectories.
\end{itemize}

\section{Related Work}
\label{sec:related_work}

Research in Multimodal Emotion Recognition (MER) has evolved from laboratory settings to in-the-wild and synthetic data, yet a perfect benchmark remains elusive. Early lab-controlled datasets like IEMOCAP \citep{busso2008iemocap} offer clean signals but lack ecological validity due to scripted interactions \citep{dhall2013emotiw}. To address this, in-the-wild datasets such as MELD \citep{poria2019meld} and CMU-MOSEI \citep{zadeh2018mosei} were introduced; however, they are often plagued by severe modality noise (e.g., canned laughter, shot transitions) and fragmented timelines that disrupt the modeling of emotion dynamics. More recently, LLM-generated datasets \citep{cheng2024emotionllama} have attempted to scale up annotation but frequently suffer from hallucinated labels and a lack of rigorous human verification \citep{ji2023selfreflection}. 

Consequently, the field faces a {``data quality trilemma''}: existing benchmarks are either ecologically invalid, excessively noisy, or unreliable. EmoS is designed to resolve this trilemma by strictly filtering for signal clarity and introducing human-verified streaming monologues. Due to space constraints, we provide a comprehensive review of existing datasets and their specific limitations in Appendix~\ref{app:related_work}.

\begin{figure*}[t]
    \centering
    \includegraphics[width=\textwidth]{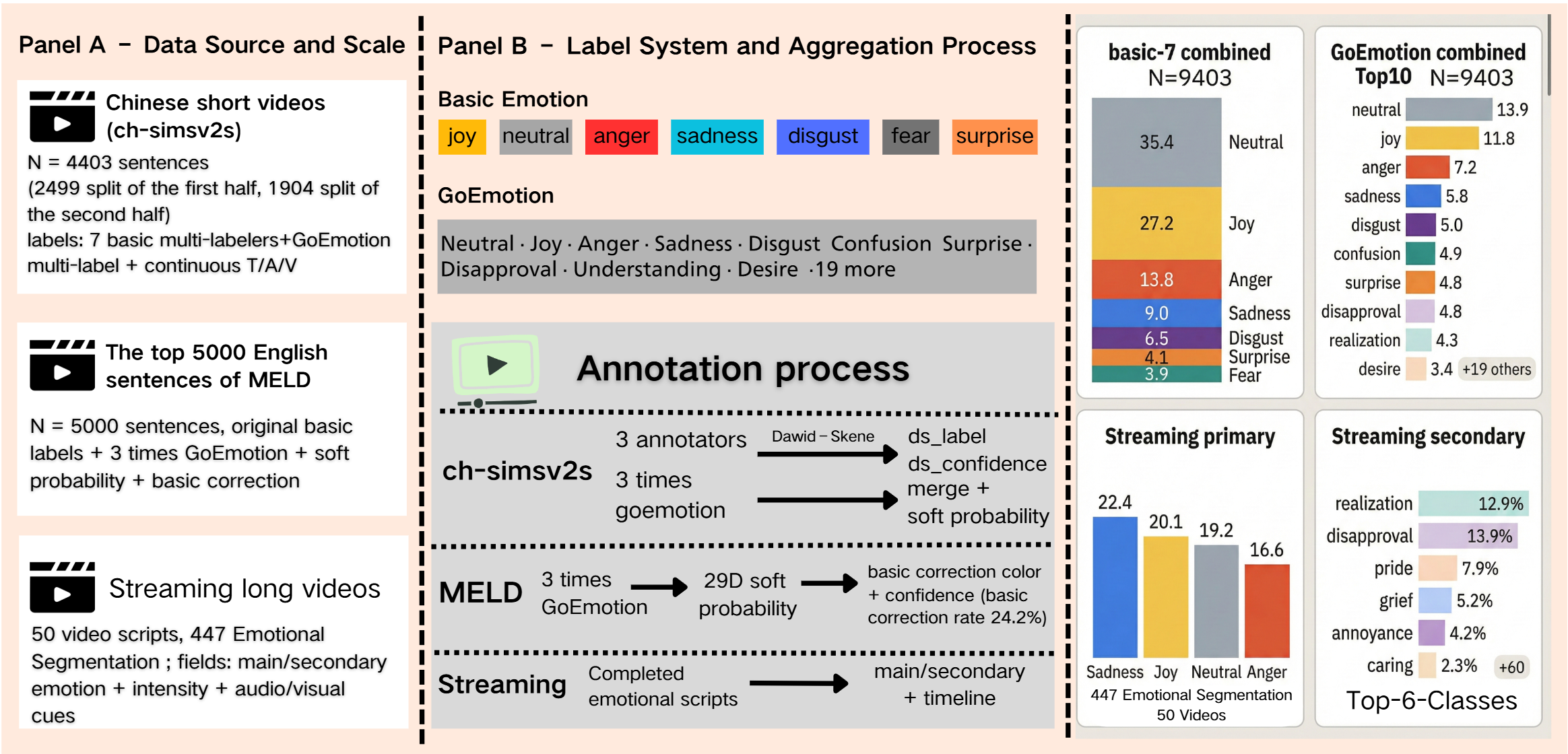}
    \caption{The basic information and processing procedures of our dataset}
    \label{fig:placeholder-1}
\end{figure*}

\section{Dataset Construction}
In the context of Multimodal Emotion Recognition (MER), the primary goal is to recognize and understand emotional expressions from multimodal data, such as speech, facial expressions, and text. Traditional MER tasks focus on recognizing emotions from short, static data slices (e.g., individual sentences or clips), without considering the temporal evolution of emotions. However, in \enquote{Streaming MER}, the task evolves to incorporate temporal dynamics, where the model must recognize and track emotional transitions across continuous streams of data (e.g., monologues, conversations). 
To address the needs of the MER task, we proposed the EmoS dataset by referring to the labeling process shown in Figure~\ref{fig:placeholder-1}, which illustrates the basic information and processing procedures of our dataset.

\subsection{Data Collection}
Our dataset, EmoS, consists of three independent subsets. It integrates carefully selected segments from MELD (English) and CH-SIMS v2 (Chinese), and is further complemented by our newly collected long-form monologue subset. The original MELD dataset \citep{poria2019meld} is derived from the sitcom Friends. It contains a large number of very short utterances (e.g., ``yes'', ``okay'') and is heavily contaminated by canned laughter. While MELD is suitable for modeling conversational context, these characteristics introduce substantial noise for fine-grained emotion recognition, as such utterances often lack standalone acoustic or visual cues. To address this issue, we perform strict filtering on MELD by removing segments shorter than one second, as well as low signal-to-noise segments caused by laughter tracks or severe visual occlusions (e.g., shot transitions that fail to maintain focus on the speaker). This yields a core subset of 5,000 high-quality samples, termed \textbf{MELD-Core}. Meanwhile, we incorporate the full CH-SIMS v2 dataset \citep{liu2022chsims2}, which provides high-quality multimodal alignment and applies video cropping for a portion of the samples to capture the speaker's visual information. This ensures that the visual focus remains on the speaker's facial expressions, offering a robust benchmark for Chinese multimodal emotion analysis.

Existing datasets often rely on sentence-level slicing or short multi-party dialogue snippets, which breaks the temporal continuity of emotional evolution. To capture dynamics such as buildup, transition, and climax, we collect 50 continuous monologues (about 2 hours total) from movies and TV dramas (e.g., The Legend of 1900 and Hei Bing). We then parse them into roughly 700 consecutive sentences, preserving the temporal coherence needed to model emotion dynamics.

\subsection{Data Annotation and Quality Analysis}
Given the limitations of current Multimodal LLMs in zero-shot emotion recognition (we tested Gemini-3 and Qwen-3-omni-flash on a pilot set, and both achieved below 70\% accuracy on the Basic-7 classification task), we adopt a strict human annotation pipeline. This section details our two-level taxonomy, quality control mechanisms, and modeling of annotator subjectivity.

\subsubsection{Annotation Protocol and Taxonomy}
To balance standardization and semantic richness, our annotation framework operates at two granularities:

\paragraph{Basic-7 (Discrete Categories).}
We follow the classic Ekman-style seven-way taxonomy (Anger, Joy, Sadness, Fear, Disgust, Surprise, Neutral). This scheme is widely used in multimodal research and provides a robust benchmark for evaluating basic discriminative capability \cite{ekman1992argument}.

\paragraph{GoEmotions-28 (Fine-grained Multi-label).}
To capture subtle affective distinctions (e.g., admiration, remorse, confusion), we adopt the 28-category fine-grained taxonomy from GoEmotions \cite{demszky-etal-2020-goemotions}. This enables further assessment of a model's capability in handling semantic proximity and complex emotion understanding.

Each sample is annotated by three independent annotators. For the CH-SIMS v2 and Streaming subsets, annotators provide both a single Basic-7 label and multi-label GoEmotions annotations. For MELD, we integrate the original Basic labels, additionally collect GoEmotions multi-label annotations, and perform cross-taxonomy consistency checks.

\subsubsection{Label Aggregation and Quality Control}
To reduce individual bias and estimate the reliability of Ground Truth \citep{whitehill2009glad,raykar2010crowds}, we implement a multi-stage aggregation and cleaning pipeline.

\paragraph{Basic-7 Aggregation (Dawid--Skene).}
For the single-label task, we use the Dawid--Skene (DS) \cite{dawid1979maximum} algorithm to estimate the inferred label $y_{ds}$ and its posterior confidence $c_{ds}$.

Specifically, the DS algorithm employs an Expectation-Maximization (EM) framework to jointly estimate the latent class priors and the confusion matrix (reliability) of each annotator. Upon convergence, the algorithm yields a posterior probability distribution over the class set $K$ for each sample $i$. The confidence score $c_{ds}^{(i)}$ is defined as the maximum value of this posterior distribution:

\begin{equation}
    c_{ds}^{(i)} = \max_{k \in K} P(y_i = k \mid x_{i}^{(1)}, x_{i}^{(2)}, x_{i}^{(3)})
\end{equation}

\noindent where $x_{i}^{(m)}$ denotes the label provided by the $m$-th annotator. This metric reflects the model's certainty in the inferred label after weighting the reliability of different annotators.
Based on $c_{ds}$ and annotator agreement, we divide the dataset into three quality tiers (statistics are shown in Table~\ref{tab:chsims_quality}):
\begin{itemize}
    \item \textbf{High-Quality (76.6\%):} all three annotators agree (unique\_labels=1) or $c_{ds} \ge 0.9$.
    \item \textbf{Medium-Quality (14.0\%):} $0.8 \le c_{ds} < 0.9$, typically involving mildly ambiguous boundaries (e.g., Sadness vs.\ Neutral).
    \item \textbf{Low-Quality / Hard (9.4\%):} $c_{ds} < 0.8$, often where all three annotators assign different labels. These samples reflect inherent subjectivity and uncertainty in human emotion perception.
\end{itemize}

\begin{table}[t]
\centering
\footnotesize
\setlength{\tabcolsep}{4pt}
\begin{tabular}{lccc}
\toprule
\textbf{Metric} & \textbf{First 2,499} & \textbf{Latter 1,904} & \textbf{Total} \\
\midrule
N & 2,499 & 1,904 & 4,403 \\
\midrule
High ($\ge$0.9) & 1,702 & 1,670 & 3,372 \\
Medium (0.8--0.9) & 496 & 122 & 618 \\
Low ($<$0.8) & 301 & 112 & 413 \\
\midrule
unique = 1 & 1,097 & 904 & 2,001 \\
unique = 2 & 1,119 & 917 & 2,036 \\
unique = 3 & 283 & 83 & 366 \\
\bottomrule
\end{tabular}
\caption{CH-SIMS v2 basic-7 classes quality tiers and unique-label counts.}
\label{tab:chsims_quality}
\end{table}

\paragraph{GoEmotions Soft-Label Modeling.}
Given the polysemy of fine-grained emotions, we avoid hard-label voting and instead model Ground Truth as a soft probability distribution. For each class $k$ among $K=29$ categories, we compute
\[
p_k = \frac{n_k}{3}, \quad p_k \in \{0, 0.33, 0.66, 1.0\},
\]
where $n_k$ is the number of annotators who select class $k$. This probabilistic representation preserves annotator disagreement and supports label distribution learning.

\begin{table}[t]
\centering
\footnotesize
\setlength{\tabcolsep}{4pt}
\begin{tabular}{lcc}
\toprule
\textbf{Metric} & \textbf{First 2,499} & \textbf{Latter 1,904} \\
\midrule
N & 2,499 & 1,904 \\
\midrule
Annotator label counts & 1.84 / 1.42 / 1.41 & 1.93 / 1.25 / 1.51 \\
Union mean & 3.68 & 3.60 \\
Union $\ge$ 4 & 1,458 & 1,521 \\
Exact match (\%) & 7.6\% & 24.4\% \\
\bottomrule
\end{tabular}
\caption{CH-SIMS v2 GoEmotions multi-label statistics.}
\label{tab:chsims_multilabel}
\end{table}

\paragraph{Basic Label Correction on MELD.}
To account for label noise in MELD, we re-aggregate Basic-7 labels and measure the correction confidence and change rate. Furthermore, we conducted a manual audit of the corrected labels in MELD. We observed that in addition to corrections arising from subjective annotator disagreements, a small fraction of the revisions addressed inherent annotation errors in the original dataset. Specifically, due to the massive scale of MELD, certain clips were duplicated but assigned inconsistent emotion labels. These discrepancies have been fully rectified in our processed subset.
Table~\ref{tab:meld_correction} summarizes the correction statistics and the corresponding GoEmotions multi-label properties.

\begin{table}
\centering
\footnotesize
\setlength{\tabcolsep}{6pt}
\begin{tabular}{lc}
\toprule
\textbf{Metric} & \textbf{Value} \\
\midrule
Basic corrected confidence $\ge$ 0.7 & 1,993 \\
Basic corrected confidence 0.5--0.7 & 2,101 \\
Basic corrected confidence $<$ 0.5 & 401 \\
Basic labels changed (basic\_changed) & 1,031 (24.2\%) \\
\midrule
GoEmotions union mean & 2.70 \\
Union $\ge$ 4 & 979 \\
Union $\ge$ 5 & 502 \\
Exact match (3 annotators) & 686 \\
\bottomrule
\end{tabular}
\caption{MELD-Core basic correction and multi-label statistics (Excluding "unsure").}
\label{tab:meld_correction}
\end{table}

\begin{table*}[t]
\centering
\small
\setlength{\tabcolsep}{3pt}
\renewcommand{\arraystretch}{1.06}

{\hyphenpenalty=10000\exhyphenpenalty=10000 
\begin{tabularx}{\textwidth}{@{}
p{2.6cm}  
p{1.1cm}  
c         
>{\raggedright\arraybackslash}X 
c         
c         
@{}}
\toprule
\textbf{Dataset / Split} & \textbf{Annotator} & $\boldsymbol{k_a}$ &
\textbf{Dominant emotion tendencies} & \textbf{Pairwise Jaccard} & \textbf{Exact match} \\
\midrule

\multirow{3}{*}{\makecell[l]{CH-SIMS v2\\First 2,499}}
& A1  & 1.84 & Joy, Disgust, Sadness, Anger, Desire, Realization (fine-grained)
& \multirow{3}{*}{0.34 -- 0.38} & \multirow{3}{*}{$\sim$8\%} \\
& A2 & 1.42 & Neutral / Unsure dominant (conservative) &  &  \\
& A3  & 1.41 & Disapproval, Confusion, Caring (cognitive/social) &  &  \\
\midrule

\multirow{3}{*}{\makecell[l]{CH-SIMS v2\\Latter 1,904}}
& A4  & $\sim$1.93 & Joy, Optimism, Caring, Excitement + Disapproval/Annoyance (expansive)
& \multirow{3}{*}{0.41 -- 0.45} & \multirow{3}{*}{$\sim$24\%} \\
& A5 & $\sim$1.25 & Near-single-label; Joy / Neutral / Anger / Sadness &  &  \\
& A6 & $\sim$1.51 & Joy, Optimism, Anger, Disappointment (bi-label) &  &  \\
\midrule

\multirow{3}{*}{\makecell[l]{MELD\\First 3,000}}
& A7  & 1.61 & Joy, Realization, Neutral + Anger / Disgust
& \multirow{3}{*}{0.389 -- 0.445} & \multirow{3}{*}{$\sim$31\%} \\
& A8 & $\sim$1.40 & Neutral dominant (conservative) &  &  \\
& A9  & $\sim$1.40 & Surprise, Confusion, Realization, Disapproval (Neutral $\sim$25\%) &  &  \\
\midrule

\multirow{3}{*}{\makecell[l]{MELD\\Latter 2,000}}
& A10  & 2.31 & Joy / Neutral + Anger / Disgust (most expansive) 
& \multirow{3}{*}{0.369 -- 0.401} & \multirow{3}{*}{$\sim$9\%} \\
& A11 & $\sim$1.50 & Neutral dominant &  &  \\
& A12  & $\sim$1.50 & Joy vs.\ Disapproval (Neutral $\sim$25\%) &  &  \\
\bottomrule
\end{tabularx}}

\caption{Annotator style summary on CH-SIMS v2 and MELD. We report per-annotator label cardinality ($k_a$) and dominant label usage. Split-level agreement is measured by pairwise Jaccard (range over three annotator pairs) and exact match (Representing the proportion agreed upon by the three annotators)}
\label{tab:annotator_style_summary}
\end{table*}

\paragraph {Streaming Monologue Subset Annotation.}
The annotation of the Streaming Monologue subset follows the same core multi-annotator pipeline as the static datasets, but it necessitates a more rigorous temporal segmentation protocol. We perform a coarse temporal segmentation based on pauses occurring in the video. Subsequently, the three annotators discuss and annotate through a meeting-based approach to collectively determine the precise segment boundaries and the overarching direction of emotion evolution, retaining only high-confidence samples. To enrich the contextual metadata of this subset, we leverage Large Language Models (LLMs) as an auxiliary tool to draft initial descriptions for emotion causes, which are subsequently verified and corrected as necessary.

\subsubsection{Annotator Modeling and Stylistic Analysis}
\begin{figure}[h]
    \centering
    \includegraphics[width=0.7\linewidth]{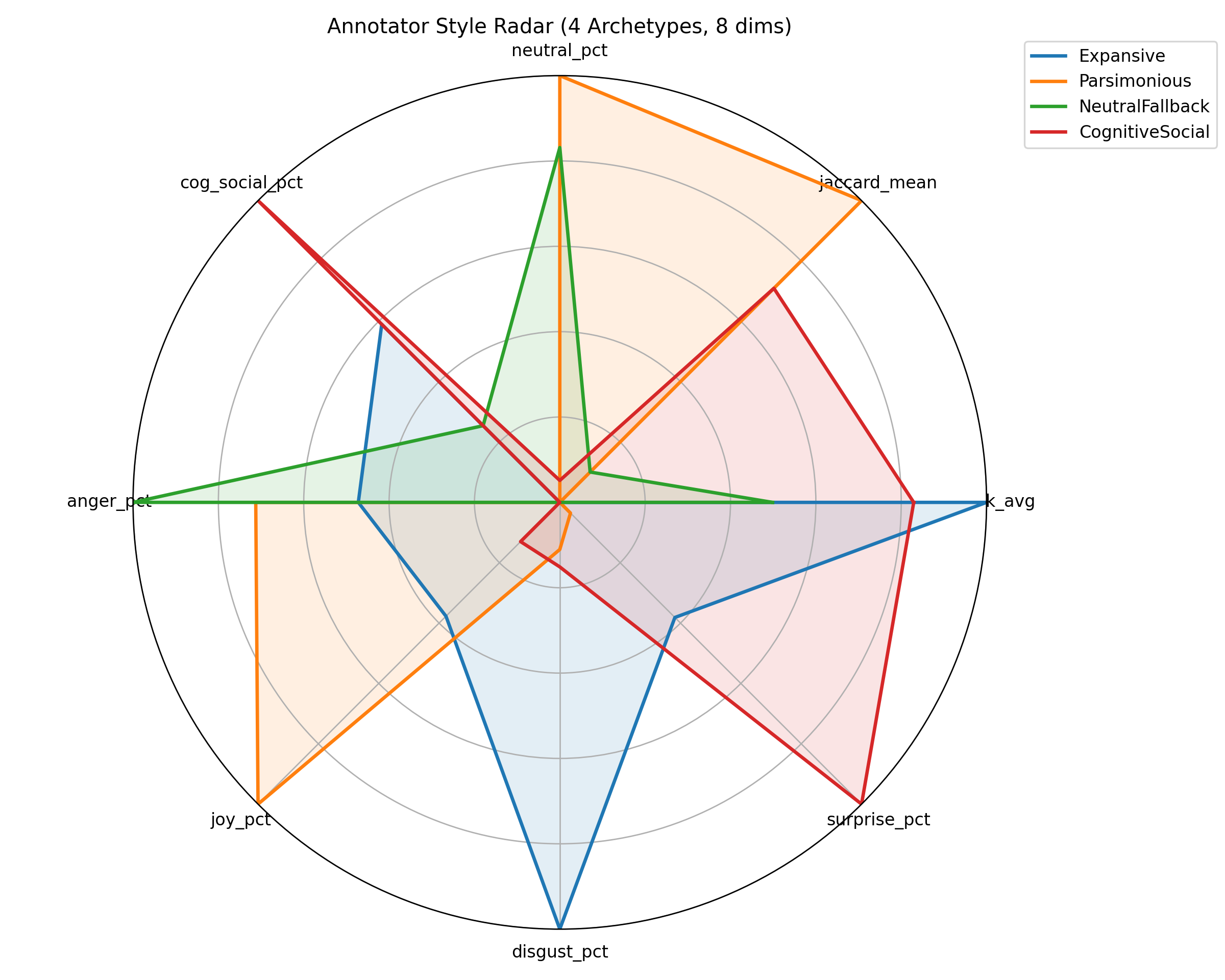}
    \caption{The performance of four annotator styles across eight representative dimensions.}
    \label{fig:ann}
\end{figure}

A key finding is that annotator styles significantly affect annotation consistency \citep{fleiss1971kappa,krippendorff2011alpha}. From GoEmotions annotation statistics (Table~\ref{tab:annotator_style_summary}), we observe that cross-subset differences in agreement are not solely driven by data noise, but also by differences in annotation style.
We characterize annotator behavior using average label cardinality (Label Cardinality, $k_a$) and semantic preference, and summarize four typical styles, as shown in (Figure.~\ref{fig:ann}) :

\paragraph{Expansive Style.}
$k_a$ is significantly above the group median. Such annotators tend to decompose complex emotions into multiple components (e.g., decomposing ``remorse'' into Sadness + Disappointment + Remorse), leading to a large label union per sample (Union $\ge 4$).

\paragraph{Parsimonious Style.}
$k_a$ is below the median. They focus on the dominant emotion, often yielding higher Exact Match agreement, but potentially missing secondary affective nuances.

\paragraph{Safety-Seeking / Neutral-Fallback.}
These annotators assign Neutral or Unsure at a notably higher rate, reflecting a conservative strategy under ambiguity to reduce the risk of mistakes.

\paragraph{Socio-Cognitive Sensitive.}
Such annotators are more sensitive to cognitive labels (e.g., Confusion, Realization) and social-evaluative labels (e.g., Caring, Disapproval). This pattern is especially evident in dialogue data (MELD), suggesting deeper attention to interpersonal cues.

This typology does not assume any style is ``more correct''; rather, it explains agreement differences. When an annotation group mixes expansive and conservative styles, the label union and disagreement (low Exact Match, low Jaccard) increase substantially. In contrast, when conservative annotators share similar decision rules, pairwise agreement and three-way agreement can remain high even with one expansive annotator present. Overall, this suggests that different populations exhibit different emotion perception tendencies; we argue that a general-purpose model should be evaluated and trained on the most broadly representative subset with maximal coverage and generality.

\subsection{Statistical Analysis}
\subsubsection{Dataset Statistics and Distributions}
The final dataset contains 9,403 samples and additional streaming data. We adopt the Dawid--Skene aggregation method, along with additional post-processing procedures, to ensure label reliability; details are provided in a later section.

\paragraph{Basic Emotion Distribution.}
As shown in Table~\ref{tab:basic_dist}, Neutral (35.4\%) and Joy (27.2\%) are relatively balanced in the combined dataset. Notably, MELD exhibits a strong neutral bias (46.4\%), reflecting the prevalence of functional utterances in everyday conversations, while CH-SIMS v2 contributes a substantial portion of positive emotion samples. Negative emotions (Anger, Sadness) remain stably represented in both languages.
\begin{table}[t]
  \centering
  \setlength{\tabcolsep}{2pt}
  \begin{tabular}{lccc}
    \toprule
    \textbf{Emotion} & \textbf{CH-SIMS v2} & \textbf{MELD-Core} & \textbf{Combined} \\
    \midrule
    Neutral  & 27.9\% & 46.4\% & 35.4\% \\
    Joy      & 31.2\% & 21.4\% & 27.2\% \\
    Anger    & 14.0\% & 13.4\% & 13.8\% \\
    Sadness  & 10.4\% & 6.9\%  & 9.0\%  \\
    Disgust  & 8.8\%  & 3.1\%  & 6.5\%  \\
    Surprise & 3.3\%  & 5.4\%  & 4.1\%  \\
    Fear     & 4.3\%  & 3.4\%  & 3.9\%  \\
    \midrule
    \textbf {Total} & 4,403 & 5,000 & 9,403 \\
    \bottomrule
  \end{tabular}
  \caption{Distribution of Basic Emotions across dataset subsets (percentages).
Percentages are computed over \textbf{non-Unsure} samples; the total row reports the \textbf{total number of samples} (including Unsure).
We additionally introduce an Unsure category for both the Basic and GoEmotions taxonomies to allow annotators to abstain under genuine ambiguity; Unsure labels are retained in the raw annotation records for potential future re-auditing.}
  \label{tab:basic_dist}
\end{table}

\paragraph{Fine-grained (GoEmotions) Analysis.}
The label distribution follows a long-tail pattern. In the Chinese subset, cognition-related labels such as ``Disapproval'' and ``Optimism'' are more prominent, whereas the English subset tends to favor labels like ``Surprise'' and ``Realization'', indicating a higher reliance on contextual inference. The dataset exhibits a pronounced long-tail distribution: labels such as Grief, Relief, and Embarrassment are extremely rare, with Grief accounting for only(0.1\%) in MELD. In addition, complex social emotions like Remorse and Gratitude generally occur at frequencies below 1.0\%, Due to space limitations, detailed data tables can be found in the Appendix~\ref{sec:Additional dataset statistics information}.

\paragraph{Dynamics in Streaming Data.}
The streaming subset is characterized by high emotional volatility. Across the 50 clips, we identify 401 emotion turning points, averaging 8.0 transitions per clip (see Table~\ref{tab:streaming_stats}). Overall, the two emotion taxonomies exhibit highly consistent yet hierarchically differentiated structural patterns in the streaming subset. Neutral emotion dominates in both schemes (Basic-7: 32.8\%; GoEmotion: 31.6\%), indicating that emotional states in continuous contexts tend to fluctuate around an affective ``baseline'' rather than remaining in highly activated states, which provides a buffer for frequent emotional shifts. Meanwhile, negative emotions are both prevalent and highly differentiated: in Basic-7, they are mainly reflected by Anger, Disgust, and Sadness, whereas GoEmotion further decomposes similar negative experiences into multiple high-frequency categories such as anger, disgust, disapproval, annoyance, and confusion. This coarse-to-fine correspondence suggests that negative affect in streaming contexts is not characterized by isolated emotional outbursts, but by the alternation of semantically related yet distinguishable emotional states. In contrast, positive emotion (Joy) maintains a moderately high but non-dominant proportion in both taxonomies (Basic-7: 19.6\%; GoEmotion: 23.7\%), implying that positive experiences tend to appear as brief insertions rather than sustained states. Collectively, these patterns form a dynamic emotional landscape characterized by a neutral background and high-frequency transitions between positive and negative states.
Due to the limitation of space, we have placed more statistical results in the Appendix~\ref{sec:Additional dataset statistics information}.

\begin{table}[h]
\centering
\small
\begin{tabular}{lr}
\toprule
\textbf{Metric} & \textbf{Value} \\
\midrule
Number of Segments & 50 \\
Total Turning Points & 401 \\
Avg. Turning Points / Segment & 8.0 \\
Max Turning Points (Sample) & 25 (Any Given Sunday) \\
\bottomrule
\end{tabular}
\caption{Statistics of the \textbf{Streaming Subset}. High frequency of turning points indicates strong emotional dynamics.}
\label{tab:streaming_stats}
\end{table}

  \begin{table*}[t]
  \centering
  \small
  \setlength{\tabcolsep}{5pt}
  \begin{tabular}{lccccc}
  \hline
  \textbf{Metric} & \textbf{Gemini-3} & \textbf{Qwen-3} & \textbf{Qwen-3 (FT)} & \textbf{Qwen-2.5} & \textbf{EmotionLLaMA} \\
  \hline
  Accuracy        & 0.611 & 0.580 & \textbf{0.703 }& 0.571 & 0.369 \\
  Macro-F1        & 0.528 & 0.517 & \textbf{0.628 }& 0.419 & 0.289 \\
  Weighted-F1     & 0.605 & 0.582 & \textbf{0.681} & 0.549 & 0.368 \\
  \hline
  Anger (P)       & 0.669 & \textbf{0.708} & 0.670 & 0.648 & 0.512 \\
  Anger (R)       & \textbf{0.770} & 0.680 & 0.759 & 0.648 & 0.536 \\
  Anger (F1)      & \textbf{0.716} & 0.694 & 0.712 & 0.648 & 0.524 \\
  \hline
  Disgust (P)     & 0.217 & 0.700 & \textbf{0.819 }& 0.444 & 0.000 \\
  Disgust (R)     & 0.143 & 0.194 & \textbf{0.704 }& 0.133 & 0.000 \\
  Disgust (F1)    & 0.172 & 0.304 & \textbf{0.757} & 0.205 & 0.000 \\
  \hline
  Fear (P)        & 0.383 & 0.449 & 0.720 & \textbf{1.000} & 0.375 \\
  Fear (R)        & \textbf{0.590} & 0.550 & 0.571 & 0.026 & 0.050 \\
  Fear (F1)       & 0.465 & 0.494 & \textbf{0.637} & 0.051 & 0.088 \\
  \hline
  Joy (P)         & 0.678 & 0.551 & 0.611 & \textbf{0.836} & 0.664 \\
  Joy (R)         & \textbf{0.802} & 0.787 & 0.533 & 0.298 & 0.393 \\
  Joy (F1)        & \textbf{0.735} & 0.648 & 0.569 & 0.439 & 0.494 \\
  \hline
  Neutral (P)     & \textbf{0.777} & 0.756 & 0.760 & 0.584 & 0.517 \\
  Neutral (R)     & 0.422 & 0.429 & \textbf{0.830} & 0.865 & 0.322 \\
  Neutral (F1)    & 0.547 & 0.547 & \textbf{0.793}& 0.697 & 0.397 \\
  \hline
  Sadness (P)     & 0.533 & 0.620 & 0.560 & \textbf{0.689} & 0.176 \\
  Sadness (R)     & \textbf{0.721} & 0.647 & 0.670 & 0.636 & 0.679 \\
  Sadness (F1)    & 0.613 & {0.633} & 0.610 & \textbf{0.661} & 0.280 \\
  \hline
  Surprise (P)    & 0.309 & 0.209 & 0.209 & \textbf{1.000} & 0.157 \\
  Surprise (R)    & \textbf{0.808} & 0.519 & 0.650 & 0.130 & 0.531 \\
  Surprise (F1)   & \textbf{0.447} & 0.298 & 0.316 & 0.231 & 0.242 \\
  \hline
  \end{tabular}
  \caption{Emotion classification results on 800 high-confidence samples.
  Class distribution (\%): Anger 15.5, Disgust 4.4, Fear 5.1, Joy 26.0, Neutral 37.1, Sadness 8.7, Surprise 3.3.}
  \label{tab:emotion_results_transposed}
  \end{table*}

\section{Experiments}

\subsection{Basic Emotion Recognition Capability}
To systematically evaluate the emotion recognition capabilities of current multimodal large language models (MLLMs), we compared a set of representative multimodal models. The closed-source model is Gemini-3 \citep{gemini_team2023gemini,gemini3pro_modelcard_2025}. The open-source baselines include Qwen-2.5-Omni-7B (hereafter abbreviated as Qwen-2.5), Qwen3-omni-flash (denoted as Qwen-3 \citep{qwen25omni_2025,qwen3omni_2025} in Table~\ref{tab:emotion_results_transposed}), and EmotionLLaMA \citep{cheng2024emotionllama}. In addition, we report a task-adapted fine-tuned variant of Qwen-3, namely \textbf{Qwen-3 (FT)}, to quantify the gain brought by task adaptation (we attempted to include AffectGPT \citep{lian2025affectgpt}, but its official implementation could not be reproduced on local devices due to environment issues). Due to space constraints, the specific settings for finetuning are presented in Appendix~\ref{sec:implementation_details}.

Except for Qwen-3 (FT), all models were evaluated under the zero-shot setting. During prompt engineering, we observed that some models could not reliably follow the instruction of 28-class GoEmotions, often collapsing to 7 classes or producing invalid labels. Therefore, to ensure cross-model comparability, we adopted the Basic-7 label space. The test set was further split into a high-confidence subset ($N=800$) for reporting the main results, and a low-confidence subset ($N=400$) for additional analyses such as model self-awareness and confidence calibration. Given the imbalanced class distribution (Neutral 37.1\%, Joy 26.0\%, while Surprise/Disgust/Fear are each below 6\%), we report Accuracy, Macro-F1, and Weighted-F1. Results on the high-confidence subset are shown in Table~\ref{tab:emotion_results_transposed}. For model predictions that failed to follow the instruction and thus fell outside the predefined set, we uniformly counted them as incorrect samples.

Overall, models are more stable on frequent classes, while their performance diverges more substantially on minority classes. This is also reflected by the fact that most systems achieve higher Weighted-F1 than Macro-F1. Under the zero-shot evaluation, Gemini-3 achieves the best overall performance (Acc 0.611 / Macro-F1 0.528 / Weighted-F1 0.605). The Qwen family shows similar Accuracy and Weighted-F1 (e.g., Qwen-3: 0.580/0.582; Qwen-2.5: 0.571/0.549), but a lower Macro-F1, indicating weaker robustness on long-tail categories. EmotionLLaMA performs the worst overall (Acc 0.369 / Macro-F1 0.289 / Weighted-F1 0.368), suggesting limited reliability on general emotion classification tasks.
Task-adaptive fine-tuning can substantially improve class-balanced performance. Qwen-3 (FT) increases Accuracy from 0.580 to 0.703 and Macro-F1 from 0.517 to 0.628 (Weighted-F1: 0.582$\rightarrow$0.681), and yields notable gains on minority classes (e.g., the F1 score of Disgust reaches 0.757, whereas zero-shot Qwen-3 attains only 0.304). At the class level, Joy and Neutral are generally easier to recognize (e.g., Gemini-3 achieves Joy F1 of 0.735; Qwen-3 (FT) achieves Neutral F1 of 0.793), while low-frequency negative emotions remain the primary bottleneck.

We identify two typical failure modes: (1) conservative prediction, where minority classes exhibit very high precision but extremely low recall (e.g., Qwen-2.5 on Fear with P=1.000 and R=0.026; on Surprise with P=1.000 and R=0.130); and (2) over-prediction, where recall is improved at the expense of precision, leading to unstable outputs (e.g., EmotionLLaMA fails completely on Disgust and shows a ``high-recall, low-precision'' pattern on Sadness). In sum, Table~\ref{tab:emotion_results_transposed} indicates that zero-shot emotion recognition in current MLLMs is mainly constrained by instruction-following stability and long-tail emotion recognition capability, while fine-tuning provides a direct path toward more reliable and more class-balanced emotion understanding.

\subsection{Fine-grained Emotion Recognition Capability (GoEmotions)}

\begin{figure}[h]
    \centering
    \includegraphics[width=1\linewidth]{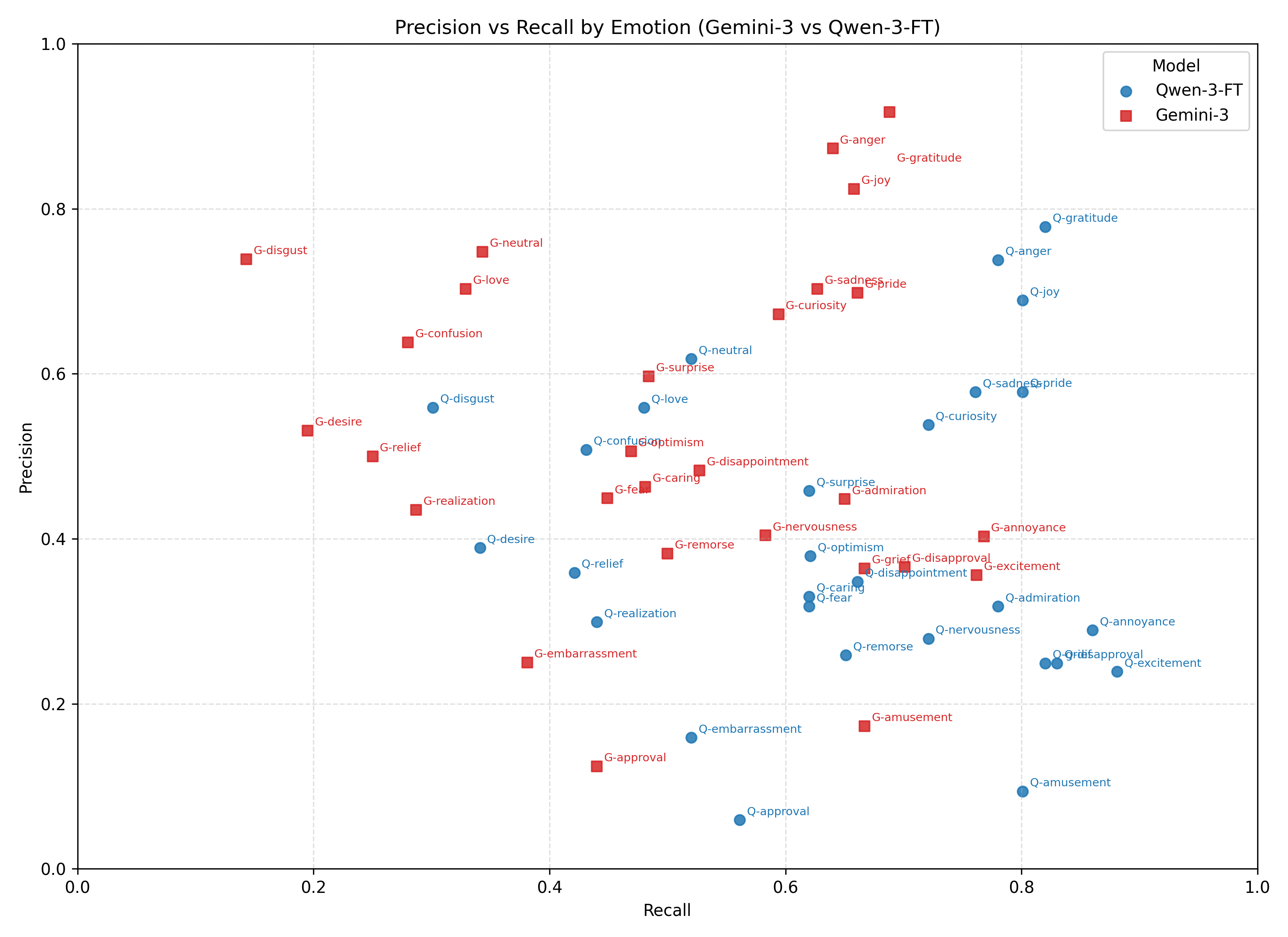}
    \caption{Precision vs. Recall per emotion label. The comparison highlights distinct preference behaviors}
    \label{fig:placeholder}
\end{figure}

We conducted a systematic comparison of the multi-label emotion classification performance of Gemini-3 and Qwen-3 (Fine-tuned). The detailed results are presented in the Appendix~\ref{sec:Detailed results of multi-classification}. Qwen-3 (Fine-tuned) exhibited higher overall recall, which was particularly pronounced for fine-grained cognitive and social emotion labels; however, this improvement was accompanied by a systematic decline in precision, leading to an increase in false positives. In contrast, Gemini-3 maintained higher and more stable precision, adopting a relatively conservative prediction strategy, but its recall on long-tail labels remained insufficient. This divergence reflects different preferences in the precision--recall trade-off: Qwen-3 (Fine-tuned) tends to expand label coverage, whereas Gemini-3 places greater emphasis on prediction reliability and false-positive control, at the expense of broader label coverage.

\subsection{Confidence Calibration and Robustness Analysis}

To examine whether the models are able to ``recognize their own uncertainty,'' we compare their output confidence scores on the high-confidence and low-confidence subsets.
As shown in Table~\ref{tab:confidence_calibration}, Gemini 3 and Qwen 3 Flash exhibit a decrease in average confidence on low-confidence samples (from 0.817 to 0.804 and from 0.873 to 0.849, respectively), indicating that they can, to some extent, perceive input uncertainty.
In contrast, Qwen2.5, constrained by its parameter scale, lacks the ability to effectively discriminate confidence levels, showing nearly identical scores across the two subsets (0.850 vs. 0.852).
Qwen3 (FT), while achieving the highest overall confidence, also exhibits the largest confidence drop ($\Delta = -0.032$).

\subsection{Evaluation on Streaming Data}
To evaluate continuous dynamics, we designed a Streaming Utterance-Level Annotation task where models must process videos to summarize emotional trajectories and annotate the main speaker's utterances using our dual-layer taxonomy. Detailed contents can be referred to Appendix~\ref{sec:prompt}. Given that Gemini-3 is currently the unique model capable of processing ultra-long multimodal context windows, our evaluation on the Streaming subset is exclusively conducted on this architecture.
We observe that as the context length expands, the model's robust text understanding capabilities significantly enhance its grasp of the narrative flow and emotional progression.
Gemini-3 demonstrates exceptional sensitivity to dynamic changes, achieving an accuracy of 82\% in predicting emotion turning points (state transitions), a task that typically challenges models limited to short-context windows. Furthermore, on the task of segmented utterance emotion classification within the continuous stream, the model attains a 71\% accuracy on the Basic-7 taxonomy and a Macro-F1 score of 0.55 on the fine-grained GoEmotions taxonomy.

\begin{table}[t]
\centering
\small
\begin{tabular}{lccc}
\toprule
Model & High Conf. & Low Conf. & $\Delta$ \\
\midrule
Gemini 3 & 0.817 & 0.804 & -0.013 \\
Qwen 3 Flash & 0.873 & 0.849 & -0.024 \\
Qwen2.5 & 0.850 & 0.852 & +0.002 \\
Qwen3(FT) & 0.987 & 0.955 & -0.032 \\
\bottomrule
\end{tabular}
\caption{Average confidence scores on the High-Confidence and Low-Confidence subsets. $\Delta$ denotes the difference (Low Conf. $-$ High Conf.).}
\label{tab:confidence_calibration}
\end{table}

\section{Conclusion}
In conclusion, we introduce EmoS, a high-precision benchmark that elevates emotion understanding from a static to a dynamic level. By rigorously filtering noise and introducing a streaming subset, EmoS resolves the limitations of fragmented datasets. Our benchmarking highlights distinct model behaviors: while Gemini-3 prioritizes precision, fine-tuned models excel in coverage, and long-context capabilities prove essential for tracking narrative shifts . Ultimately, EmoS lays the foundation for training and evaluating future MLLMs capable of mastering emotional variations.

\section*{Limitations}

While EmoS prioritizes label fidelity through a rigorous dual-layer human annotation pipeline, this focus inevitably constrains the scale of the newly introduced streaming dataset relative to massive web-crawled corpora. Specifically, the Streaming Monologue subset, comprising 50 videos, serves primarily as a high-standard evaluation benchmark rather than a large-scale training source. We are committed to iteratively updating and expanding this Streaming Monologue subset; however, this process requires time, primarily due to the limited availability of suitable movie clips and video content. Subsequent progress will be synchronously updated on our GitHub.

Furthermore, the current iteration is restricted to Chinese and English; expanding to a broader spectrum of languages remains necessary to enhance cross-cultural generalization. Moreover, the inherent long-tail distribution of real-world emotions---where categories such as \textit{Grief} account for only $0.1\%$ of samples---similarly manifests within this dataset. This indicates a critical need for the development of more robust imbalanced learning strategies in future work.

\section*{Acknowledgments}

This work was supported in part by the Science and Technology Development Fund of Macau SAR (Grant Nos. FDCT/0007/2024/AKP, EF2024-00185-FST), the UM and UMDF (Grant Nos. MYRG-GRG2024-00165-FST-UMDF, MYRG-GRG2025-00236-FST), the Tencent AI Lab Rhino-Bird Research Program (Grant No. EF2023-00151-FST), the Stanley Ho Medical Development Foundation (Grant No. SHMDF-AI/2026/001), the Macao Young Scholars Program (Grant No. AM2024015), and the National Natural Science Foundation of China (Grant No. 62266013). We thank Pengyu Chen, Zirui Chen, Yanxiu Liu, Fangfei Ren, Siqi Chen, and all the annotators for their valuable contributions to building the EmoS dataset.

\bibliography{custom}

\begin{thebibliography}{29}
\providecommand{\natexlab}[1]{#1}

\bibitem[{Aguilera et~al.(2023)Aguilera, Mellado, and Rojas}]{kanluan2023inwild_assessment}
Ana Aguilera, Diego Mellado, and Felipe Rojas. 2023.
\newblock \href {https://doi.org/10.3390/s23115184} {An assessment of in-the-wild datasets for multimodal emotion recognition}.
\newblock \emph{Sensors}, 23(11):5184.

\bibitem[{Busso et~al.(2008)Busso, Bulut, Lee, Kazemzadeh, Mower, Kim, Chang, Lee, and Narayanan}]{busso2008iemocap}
Carlos Busso, Murtaza Bulut, Chi-Chun Lee, Abe Kazemzadeh, Emily Mower, Samuel Kim, Jeannette~N Chang, Sungbok Lee, and Shrikanth~S Narayanan. 2008.
\newblock \href {https://doi.org/10.1007/s10579-008-9076-6} {{IEMOCAP}: Interactive emotional dyadic motion capture database}.
\newblock \emph{Language Resources and Evaluation}, 42(4):335--359.

\bibitem[{Cheng et~al.(2024)Cheng, Cheng, He, Sun, Wang, Lin, Lian, Peng, and Hauptmann}]{cheng2024emotionllama}
Zebang Cheng, Zhi-Qi Cheng, Jun-Yan He, Jingdong Sun, Kai Wang, Yuxiang Lin, Zheng Lian, Xiaojiang Peng, and Alexander~G. Hauptmann. 2024.
\newblock \href {https://arxiv.org/abs/2406.11161} {{Emotion-LLaMA}: Multimodal emotion recognition and reasoning with instruction tuning}.
\newblock In \emph{Advances in Neural Information Processing Systems}.

\bibitem[{Dawid and Skene(1979)}]{dawid1979maximum}
A.~P. Dawid and A.~M. Skene. 1979.
\newblock Maximum likelihood estimation of observer error-rates using the em algorithm.
\newblock \emph{Journal of the Royal Statistical Society. Series C (Applied Statistics)}, 28(1):20--28.
\newblock Original Dawid--Skene model for aggregating multiple annotator labels via EM.

\bibitem[{Demszky et~al.(2020)Demszky, Movshovitz-Attias, Ko, Cowen, Nemade, and Ravi}]{demszky-etal-2020-goemotions}
Dorottya Demszky, Dana Movshovitz-Attias, Jeongwoo Ko, Alan Cowen, Gaurav Nemade, and Sujith Ravi. 2020.
\newblock \href {https://doi.org/10.18653/v1/2020.acl-main.372} {Goemotions: A dataset of fine-grained emotions}.
\newblock In \emph{Proceedings of the 58th Annual Meeting of the Association for Computational Linguistics}, pages 4040--4054, Online. Association for Computational Linguistics.

\bibitem[{Dhall et~al.(2013)Dhall, Goecke, Joshi, Wagner, and Gedeon}]{dhall2013emotiw}
Abhinav Dhall, Roland Goecke, Jyoti Joshi, Michael Wagner, and Tom Gedeon. 2013.
\newblock \href {https://doi.org/10.1145/2522848.2531739} {The {E}motion {R}ecognition in the {W}ild challenge 2013}.
\newblock In \emph{Proceedings of the 15th ACM International Conference on Multimodal Interaction}, pages 509--515.

\bibitem[{Ekman(1992)}]{ekman1992argument}
Paul Ekman. 1992.
\newblock \href {https://doi.org/10.1080/02699939208411068} {An argument for basic emotions}.
\newblock \emph{Cognition \& Emotion}, 6(3/4):169--200.
\newblock Classic argument for discrete basic emotions (e.g., anger, joy, fear, sadness, disgust, surprise).

\bibitem[{Fleiss(1971)}]{fleiss1971kappa}
Joseph~L. Fleiss. 1971.
\newblock \href {https://doi.org/10.1037/h0031619} {Measuring nominal scale agreement among many raters}.
\newblock \emph{Psychological Bulletin}, 76(5):378--382.

\bibitem[{{Gemini Team}(2023)}]{gemini_team2023gemini}
{Gemini Team}. 2023.
\newblock \href {https://arxiv.org/abs/2312.11805} {Gemini: A family of highly capable multimodal models}.

\bibitem[{{Google}(2025)}]{gemini3pro_modelcard_2025}
{Google}. 2025.
\newblock \href {https://storage.googleapis.com/deepmind-media/Model-Cards/Gemini-3-Pro-Model-Card.pdf} {Gemini 3 pro model card}.
\newblock Model card listed on Google's official Model Cards site.

\bibitem[{Gratch et~al.(2014)Gratch, Artstein, Lucas, Stratou, Scherer, Nazarian, Wood, Boberg, DeVault, Marsella, Traum, Rizzo, and Morency}]{gratch2014daic}
Jonathan Gratch, Ron Artstein, Gale Lucas, Giota Stratou, Stefan Scherer, Angela Nazarian, Rachel Wood, Jill Boberg, David DeVault, Stacy Marsella, David Traum, Skip Rizzo, and Louis-Philippe Morency. 2014.
\newblock \href {https://aclanthology.org/L14-1421/} {The distress analysis interview corpus of human and computer interviews}.
\newblock In \emph{Proceedings of the Ninth International Conference on Language Resources and Evaluation (LREC'14)}, pages 3123--3128, Reykjavik, Iceland. European Language Resources Association (ELRA).

\bibitem[{Huang et~al.(2024)Huang, Yu, Ma, Zhong, Feng, Wang, Chen, Peng, Feng, Qin, and Liu}]{huang2023hallucination_survey}
Lei Huang, Weijiang Yu, Weitao Ma, Weihong Zhong, Zhangyin Feng, Haotian Wang, Qianglong Chen, Weihua Peng, Xiaocheng Feng, Bing Qin, and Ting Liu. 2024.
\newblock \href {https://doi.org/10.1145/3703155} {A survey on hallucination in large language models: Principles, taxonomy, challenges, and open questions}.

\bibitem[{Ji et~al.(2023)Ji, Yu, Xu, Lee, Ishii, and Fung}]{ji2023selfreflection}
Ziwei Ji, Tiezheng Yu, Yan Xu, Nayeon Lee, Etsuko Ishii, and Pascale Fung. 2023.
\newblock \href {https://doi.org/10.18653/v1/2023.findings-emnlp.123} {Towards mitigating hallucination in large language models via self-reflection}.
\newblock In \emph{Findings of the Association for Computational Linguistics: EMNLP 2023}, pages 1827--1843, Singapore. Association for Computational Linguistics.

\bibitem[{Jiang et~al.(2020)Jiang, Zong, Zheng, Tang, Xia, Lu, and Liu}]{jiang2020dfew}
Xingxun Jiang, Yuan Zong, Wenming Zheng, Chuangao Tang, Wanchuang Xia, Cheng Lu, and Jiateng Liu. 2020.
\newblock \href {https://doi.org/10.1145/3394171.3413620} {Dfew: A large-scale database for recognizing dynamic facial expressions in the wild}.
\newblock In \emph{Proceedings of the 28th {ACM} International Conference on Multimedia}, pages 2881--2889, New York, NY, USA. {ACM}.

\bibitem[{Krippendorff(2011)}]{krippendorff2011alpha}
Klaus Krippendorff. 2011.
\newblock \href {https://repository.upenn.edu/asc_papers/43/} {Computing krippendorff's alpha-reliability}.
\newblock Technical report, University of Pennsylvania.

\bibitem[{Lian et~al.(2025)Lian, Chen, Chen, Sun, Sun, Ren, Cheng, Liu, Liu, Peng, Yi, and Tao}]{lian2025affectgpt}
Zheng Lian, Haoyu Chen, Lan Chen, Haiyang Sun, Licai Sun, Yong Ren, Zebang Cheng, Bin Liu, Rui Liu, Xiaojiang Peng, Jiangyan Yi, and Jianhua Tao. 2025.
\newblock \href {https://arxiv.org/abs/2501.16566} {Affectgpt: A new dataset, model, and benchmark for emotion understanding with multimodal large language models}.
\newblock In \emph{Proceedings of the 42nd International Conference on Machine Learning}, volume 267 of \emph{Proceedings of Machine Learning Research}, pages 36993--37014.

\bibitem[{Liu et~al.(2022{\natexlab{a}})Liu, Yuan, Mao, Liang, Yang, Qiu, Cheng, Li, Xu, and Gao}]{liu2022chsims2}
Yihe Liu, Ziqi Yuan, Huisheng Mao, Zhiyun Liang, Wanqiuyue Yang, Yuanzhe Qiu, Tie Cheng, Xiaoteng Li, Hua Xu, and Kai Gao. 2022{\natexlab{a}}.
\newblock \href {https://doi.org/10.1145/3536221.3556630} {Make acoustic and visual cues matter: {CH-SIMS} v2.0 dataset and {AV-Mixup} consistent module}.
\newblock In \emph{Proceedings of the 24th International Conference on Multimodal Interaction}.

\bibitem[{Liu et~al.(2022{\natexlab{b}})Liu, Dai, Feng, Wang, Yin, Zeng, and Shan}]{liu2022mafw}
Yuanyuan Liu, Wei Dai, Chuanxu Feng, Wenbin Wang, Guanghao Yin, Jiabei Zeng, and Shiguang Shan. 2022{\natexlab{b}}.
\newblock \href {https://doi.org/10.1145/3503161.3548190} {Mafw: A large-scale, multi-modal, compound affective database for dynamic facial expression recognition in the wild}.
\newblock In \emph{Proceedings of the 30th {ACM} International Conference on Multimedia}, New York, NY, USA. {ACM}.

\bibitem[{McKeown et~al.(2012)McKeown, Valstar, Cowie, Pantic, and Schr{\"o}der}]{mckeown2012semaine}
Gary McKeown, Michel Valstar, Roddy Cowie, Maja Pantic, and Marc Schr{\"o}der. 2012.
\newblock \href {https://doi.org/10.1109/T-AFFC.2011.20} {The {SEMAINE} database: Annotated multimodal records of emotionally colored conversations between a person and a limited agent}.
\newblock \emph{IEEE Transactions on Affective Computing}, 3(1):5--17.

\bibitem[{Mollahosseini et~al.(2019)Mollahosseini, Hasani, and Mahoor}]{mollahosseini2017affectnet}
Amir Mollahosseini, Behzad Hasani, and Mohammad~H. Mahoor. 2019.
\newblock \href {https://doi.org/10.1109/TAFFC.2017.2740923} {{AffectNet}: A database for facial expression, valence, and arousal computing in the wild}.
\newblock \emph{IEEE Transactions on Affective Computing}, 10(1):18--31.

\bibitem[{Poria et~al.(2019)Poria, Hazarika, Majumder, Naik, Cambria, and Mihalcea}]{poria2019meld}
Soujanya Poria, Devamanyu Hazarika, Navonil Majumder, Gautam Naik, Erik Cambria, and Rada Mihalcea. 2019.
\newblock \href {https://doi.org/10.18653/v1/P19-1050} {{MELD}: A multimodal multi-party dataset for emotion recognition in conversations}.
\newblock In \emph{Proceedings of ACL}, pages 527--536.

\bibitem[{Raykar et~al.(2010)Raykar, Yu, Zhao, Valadez, Florin, Bogoni, and Moy}]{raykar2010crowds}
Vikas~C. Raykar, Shipeng Yu, Linda~H. Zhao, Gerardo~Hermosillo Valadez, Charles Florin, Luca Bogoni, and Linda Moy. 2010.
\newblock \href {https://jmlr.org/papers/v11/raykar10a.html} {Learning from crowds}.
\newblock \emph{Journal of Machine Learning Research}, 11:1297--1322.

\bibitem[{Ringeval et~al.(2013)Ringeval, Sonderegger, Sauer, and Lalanne}]{ringeval2013recola}
Fabien Ringeval, Andreas Sonderegger, J{\"u}rgen~S. Sauer, and Denis Lalanne. 2013.
\newblock \href {https://doi.org/10.1109/FG.2013.6553805} {Introducing the {RECOLA} multimodal corpus of remote collaborative and affective interactions}.
\newblock In \emph{2013 10th {IEEE} International Conference and Workshops on Automatic Face and Gesture Recognition ({FG})}, pages 1--8, Shanghai, China. IEEE.

\bibitem[{Shi et~al.(2025)Shi, Shu, Dong, Liu, Sesay, Li, and Hu}]{voila2025}
Yemin Shi, Yu~Shu, Siwei Dong, Guangyi Liu, Jaward Sesay, Jingwen Li, and Zhiting Hu. 2025.
\newblock \href {https://arxiv.org/abs/2505.02707} {Voila: Voice-language foundation models for real-time autonomous interaction and voice role-play}.
\newblock \emph{arXiv}.

\bibitem[{Whitehill et~al.(2009)Whitehill, Wu, Bergsma, Movellan, and Ruvolo}]{whitehill2009glad}
Jacob Whitehill, Ting-Fan Wu, Jacob Bergsma, Javier~R. Movellan, and Paul~L. Ruvolo. 2009.
\newblock \href {https://papers.nips.cc/paper_files/paper/2009/hash/0c6415c27e9e6c31f9f6c0f6a0c3d1b0-Abstract.html} {Whose vote should count more: Optimal integration of labels from labelers of unknown expertise}.
\newblock In \emph{Advances in Neural Information Processing Systems}.

\bibitem[{Xu et~al.(2025{\natexlab{a}})Xu, Guo, He, Hu, He, Bai, Chen, Wang, Fan, Dang, Zhang, Wang, Chu, and Lin}]{qwen25omni_2025}
Jin Xu, Zhifang Guo, Jinzheng He, Hangrui Hu, Ting He, Shuai Bai, Keqin Chen, Jialin Wang, Yang Fan, Kai Dang, Bin Zhang, Xiong Wang, Yunfei Chu, and Junyang Lin. 2025{\natexlab{a}}.
\newblock \href {https://arxiv.org/abs/2503.20215} {Qwen2.5-omni technical report}.
\newblock \emph{Preprint}, arXiv:2503.20215.

\bibitem[{Xu et~al.(2025{\natexlab{b}})Xu, Guo, Hu, Chu, Wang, He, Wang, Shi, He, Zhu, Lv, Wang, Guo, Wang, Ma, Zhang, Zhang, Hao, Guo, Yang, Zhang, Ma, Wei, Bai, Chen, Liu, Wang, Yang, Liu, Ren, Zheng, Men, Zhou, Yu, Yang, Yu, Zhou, and Lin}]{qwen3omni_2025}
Jin Xu, Zhifang Guo, Hangrui Hu, Yunfei Chu, Xiong Wang, Jinzheng He, Yuxuan Wang, Xian Shi, Ting He, Xinfa Zhu, Yuanjun Lv, Yongqi Wang, Dake Guo, He~Wang, Linhan Ma, Pei Zhang, Xinyu Zhang, Hongkun Hao, Zishan Guo, and 19 others. 2025{\natexlab{b}}.
\newblock \href {https://arxiv.org/abs/2509.17765} {Qwen3-omni technical report}.
\newblock \emph{Preprint}, arXiv:2509.17765.

\bibitem[{Zadeh et~al.(2016)Zadeh, Zellers, Pincus, and Morency}]{zadeh2016mosi}
Amir Zadeh, Rowan Zellers, Eli Pincus, and Louis-Philippe Morency. 2016.
\newblock \href {https://doi.org/10.1109/MIS.2016.94} {Multimodal sentiment intensity analysis in videos: Facial gestures and verbal messages}.
\newblock \emph{IEEE Intelligent Systems}, 31(6):82--88.

\bibitem[{Zadeh et~al.(2018)Zadeh, Liang, Poria, Cambria, and Morency}]{zadeh2018mosei}
AmirAli~Bagher Zadeh, Paul~Pu Liang, Soujanya Poria, Erik Cambria, and Louis-Philippe Morency. 2018.
\newblock \href {https://doi.org/10.18653/v1/P18-1208} {Multimodal language analysis in the wild: {CMU-MOSEI} dataset and interpretable dynamic fusion graph}.
\newblock In \emph{Proceedings of ACL}, pages 2236--2246.

\end{thebibliography}

\appendix
\section{Copyright and Data Usage}
This project strictly adheres to the copyright and license terms of all source datasets.

For MELD and CH-SIMS v2, we do not redistribute any original data, including but not limited to videos, audio tracks, or utterance transcripts. Instead, we only release re-annotation files in CSV format. These files contain video-level or utterance-level IDs and the corresponding newly assigned labels, and are designed to be directly compatible with the original datasets.
Users can simply replace the original annotation CSV files with our provided CSV files after obtaining the raw data from the official sources. Access to the original MELD and CH-SIMS v2 datasets must be requested and downloaded through their respective official channels, in accordance with their original licenses.

In contrast, for our self-collected streaming dataset, we hold the rights to distribute the data. Therefore, both the annotations and the corresponding video files are packaged together and publicly released via an open online repository for research purposes.

By clearly separating re-annotations of third-party datasets from our own data, we ensure full compliance with existing dataset licenses while facilitating reproducible research.

The streaming video clips used in this dataset are sourced from openly available segments on YouTube, TikTok, and Bilibili, and the original video URLs are provided in the accompanying CSV files for transparency and traceability.

\section{Finetuning Details}
\label{sec:implementation_details}

For the fine-tuning of \textbf{Qwen-3} on the EmoS dataset, we conducted instruction tuning using a mixture of high-confidence and low-confidence samples, with 80\% of the dataset being used for training and 20\% reserved for testing. To reduce computational costs while maintaining performance, we employed Low-Rank Adaptation (LoRA) with a rank of 16 and a scaling factor (alpha) of 32, setting the dropout rate to 0.0. LoRA adapters were applied to all linear layers, including the attention module's $q\_proj$, $k\_proj$, $v\_proj$, $o\_proj$, and the feed-forward network's gate, up, and down projection layers.

During training, we used the AdamW optimizer with an initial learning rate of $5 \times 10^{-5}$, paired with a cosine learning rate decay strategy and 10\% warmup steps. We set the batch size to 1 on a single H800 GPU and applied gradient accumulation over 8 steps. The training was conducted for 3 epochs under BF16 mixed precision until the loss converged.

\section{Additional dataset statistics information}
\label{sec:Additional dataset statistics information}
The following table contains additional statistical information of the dataset. Due to the length constraints of the paper, this information cannot be included in the main text. Please refer to table\ref{tab:emotion28_stats} \ref{tab:streaming_goemotions_2perrow} \ref{tab:streaming_basic_2perrow}.

\begin{table}[h]
\centering
\small
\begin{tabular}{lrrr}
\hline
\textbf{Emotion} & \textbf{CN (\%)} & \textbf{EN (\%)} & \textbf{ALL (\%)} \\
\hline
admiration      & 7.06  & 1.70  & 5.05 \\
amusement       & 6.24  & 3.23  & 5.11 \\
anger           & 16.14 & 14.43 & 15.50 \\
annoyance       & 13.26 & 8.10  & 11.33 \\
approval        & 2.92  & 3.00  & 2.95 \\
caring          & 11.90 & 5.10  & 9.35 \\
confusion       & 13.38 & 14.57 & 13.83 \\
curiosity       & 5.50  & 11.03 & 7.58 \\
desire          & 10.34 & 9.47  & 10.01 \\
disappointment  & 9.76  & 3.17  & 7.29 \\
disapproval     & 16.06 & 11.67 & 14.41 \\
disgust         & 15.46 & 14.43 & 15.07 \\
embarrassment   & 1.52  & 4.40  & 2.60 \\
excitement      & 10.26 & 4.60  & 8.14 \\
fear            & 5.34  & 4.13  & 4.89 \\
gratitude       & 3.22  & 1.23  & 2.48 \\
grief           & 1.70  & 0.20  & 1.14 \\
joy             & 28.64 & 23.03 & 26.54 \\
love            & 7.54  & 8.50  & 7.90 \\
nervousness     & 8.18  & 7.10  & 7.78 \\
neutral         & 24.04 & 45.93 & 32.25 \\
optimism        & 17.20 & 2.67  & 11.75 \\
pride           & 9.64  & 2.97  & 7.14 \\
realization     & 11.84 & 25.53 & 16.98 \\
relief          & 3.80  & 2.47  & 3.30 \\
remorse         & 3.06  & 2.50  & 2.85 \\
sadness         & 14.22 & 11.43 & 13.18 \\
surprise        & 11.14 & 14.60 & 12.44 \\
\hline
\end{tabular}
\caption{Statistics of the 28-class emotion dataset (CN/EN/ALL). Excluding "unsure"}
\label{tab:emotion28_stats}
\end{table}

\begin{table}[h]
\centering
\small
\begin{tabular}{l r | l r}
\hline
\textbf{Emotion} & \textbf{\%} & \textbf{Emotion} & \textbf{\%} \\
\hline
admiration     & 9.84  & amusement      & 6.26  \\
anger          & 9.17  & annoyance      & 12.75 \\
approval       & 7.38  & caring         & 10.51 \\
confusion      & 6.71  & curiosity      & 5.37  \\
desire         & 5.82  & disappointment & 9.17  \\
disapproval    & 21.92 & disgust        & 6.94  \\
embarrassment  & 1.34  & excitement     & 8.95  \\
fear           & 1.79  & gratitude      & 1.12  \\
grief          & 13.20 & joy            & 0.89  \\
love           & 4.03  & nervousness    & 5.37  \\
neutral        & 2.24  & optimism       & 7.83  \\
pride          & 15.44 & realization    & 22.60 \\
relief         & 4.25  & remorse        & 9.17  \\
sadness        & 3.36  & surprise       & 0.22  \\
\hline
\end{tabular}
\caption{Emotion distribution of the streaming subset (GoEmotions, 28 classes).}
\label{tab:streaming_goemotions_2perrow}
\end{table}

\begin{table}[h]
\centering
\small
\begin{tabular}{l r | l r}
\hline
\textbf{Emotion} & \textbf{\%} & \textbf{Emotion} & \textbf{\%} \\
\hline
sadness  & $\sim$24.2 & anger    & $\sim$23.0 \\
neutral  & $\sim$20.6 & joy      & $\sim$20.4 \\
disgust  & $\sim$6.7  & fear     & $\sim$3.8  \\
surprise & $\sim$1.3  &          &            \\
\hline
\end{tabular}
\caption{Emotion distribution of the streaming subset (basic 7 classes).}
\label{tab:streaming_basic_2perrow}
\end{table}

\section{Detailed results of multi-classification}
\label{sec:Detailed results of multi-classification}
 Please refer to table\ref{Detailed results of multi-classification} 
\begin{table}[h]
\centering
\small
\begin{tabular}{lcccc}
\hline
Emotion & \multicolumn{2}{c}{Gemini-3} & \multicolumn{2}{c}{Qwen-3 (FT)} \\
 & P & R & P & R \\
\hline
admiration     & 0.448 & 0.650 & 0.320 & 0.780 \\
amusement      & 0.173 & 0.667 & 0.095 & 0.800 \\
anger          & 0.873 & 0.640 & 0.740 & 0.780 \\
annoyance      & 0.403 & 0.768 & 0.290 & 0.860 \\
approval       & 0.124 & 0.440 & 0.060 & 0.560 \\
caring         & 0.463 & 0.481 & 0.330 & 0.620 \\
confusion      & 0.638 & 0.280 & 0.510 & 0.430 \\
curiosity      & 0.672 & 0.594 & 0.540 & 0.720 \\
desire         & 0.531 & 0.195 & 0.390 & 0.340 \\
disappointment & 0.483 & 0.527 & 0.350 & 0.660 \\
disapproval    & 0.366 & 0.701 & 0.250 & 0.830 \\
disgust        & 0.739 & 0.143 & 0.560 & 0.300 \\
embarrassment  & 0.250 & 0.381 & 0.160 & 0.520 \\
excitement     & 0.356 & 0.762 & 0.240 & 0.880 \\
fear           & 0.449 & 0.449 & 0.320 & 0.620 \\
gratitude      & 0.917 & 0.688 & 0.780 & 0.820 \\
grief          & 0.364 & 0.667 & 0.250 & 0.820 \\
joy            & 0.824 & 0.658 & 0.690 & 0.800 \\
love           & 0.703 & 0.329 & 0.560 & 0.480 \\
nervousness    & 0.404 & 0.583 & 0.280 & 0.720 \\
optimism       & 0.506 & 0.469 & 0.380 & 0.620 \\
pride          & 0.698 & 0.661 & 0.580 & 0.800 \\
realization    & 0.435 & 0.287 & 0.300 & 0.440 \\
relief         & 0.500 & 0.250 & 0.360 & 0.420 \\
remorse        & 0.382 & 0.500 & 0.260 & 0.650 \\
sadness        & 0.703 & 0.627 & 0.580 & 0.760 \\
surprise       & 0.597 & 0.484 & 0.460 & 0.620 \\
neutral        & 0.748 & 0.343 & 0.620 & 0.520 \\
\hline
\end{tabular}
\caption{Detailed results of multi-classification (GoEmotions, 28 classes).}
\label{Detailed results of multi-classification}
\end{table}

\section{Chinese and English distinction}
We also recorded the evaluation results for the Chinese and English subsets and have placed them in the appendix for reference. Please refer to table\ref{tab:main_results}.

\begin{table}[h]
\centering
\small
\setlength{\tabcolsep}{5pt} 
\resizebox{\columnwidth}{!}{%
\begin{tabular}{lcccc}
\toprule
\textbf{Model} & \textbf{Lang.} & \textbf{Accuracy} & \textbf{Macro-F1} & \textbf{Weighted-F1} \\
\midrule
\multirow{2}{*}{EmotionLLaMA} & En & 0.333 & 0.250 & 0.380 \\
                              & Zh & 0.512 & 0.332 & 0.489 \\
\midrule
\multirow{2}{*}{Qwen-2.5}     & En & 0.578 & 0.371 & 0.556 \\
                              & Zh & 0.564 & 0.406 & 0.541 \\
\midrule
\multirow{2}{*}{Qwen-3 (Base)}& En & 0.466 & 0.368 & 0.485 \\
                              & Zh & 0.693 & 0.624 & 0.686 \\
\midrule
\multirow{2}{*}{Gemini}       & En & 0.541 & 0.441 & 0.554 \\
                              & Zh & 0.682 & 0.597 & 0.668 \\
\midrule
\multirow{2}{*}{\textbf{Qwen-3 (FT)}} & En & 0.668 & 0.591 & 0.661 \\
                                        & Zh & 0.732 & 0.653 & 0.701 \\
\bottomrule
\end{tabular}
}
\caption{Performance comparison on high-confidence samples across different models. "En" denotes English and "Zh" denotes Chinese.}
\label{tab:main_results}
\end{table}

\section{Data Structure and Annotation Schema}
\label{sec:data_structure}

The EmoS dataset is organized into three primary directories, each catering to different modalities and annotation granularities. The detailed structure and schema definitions are provided below.

\subsection{Directory Organization}

\paragraph{1. Streaming Long-form Video (\texttt{streaming/})}
This directory contains long-form monologues. Each entry includes both the original full video and segmented clips. The data is paired with two types of annotation files:
\begin{itemize}
    \item \textbf{Interpretation JSON:} Captures the narrative flow. Keys include \texttt{summary}, \texttt{overall\_emotion\_trend} (dominant emotion + trajectory description), and a list of \texttt{utterances}. Each utterance object contains the time range, bilingual transcripts, primary/secondary emotions, multimodal cues (audio/visual), intensity, and evidence.
    \item \textbf{Timestamp JSON:} Provides strict temporal alignment. The top level includes metadata (language, title, global start/end). The \texttt{segments} list details the start/end seconds, text, and the corresponding slice filename for every sentence.
    \item \textbf{Cross-Index:} A sentence-level aggregation JSON that indexes all samples with fixed fields: language, global index, relative source path, title, timestamps, full transcript, and sub-sentence segment arrays.
\end{itemize}

\paragraph{2. Multi-Annotator Short Video (\texttt{ch-simsv2s/})}
This subset focuses on short video clips with multiple human annotators per sample. The raw index contains \texttt{video\_id}, \texttt{clip\_id}, and multimodal tags. Annotations are split into two batches, each reviewed by three annotators for both Basic-7 (single-label) and GoEmotions (multi-label).
\begin{itemize}
    \item \textbf{Basic-7 Aggregation:} Contains the three individual labels and the aggregated ground truth derived via the Dawid-Skene algorithm (\texttt{ds\_label}). It includes the aggregation confidence (\texttt{ds\_confidence}) and a \texttt{unique\_labels} field to indicate conflict levels.
    \item \textbf{GoEmotions Set:} Stores the set of labels from all three annotators, including derived metrics such as union, intersection, and label distribution summaries.
\end{itemize}

\paragraph{3. Re-annotated MELD (\texttt{newdataset-MELD/})}
A subset of MELD re-annotated to bridge the gap between simple sentiment and complex emotions. 
\begin{itemize}
    \item \textbf{GoEmotions Soft-Labels:} Contains the label sets from three annotators. It calculates the frequency/probability for each emotion (\texttt{Goemotions\_prob\_*}) and maps these soft labels to predict the best/second-best Basic-7 categories.
    \item \textbf{Basic-7 Hierarchical Table:} Provides a comparison between the original MELD emotion (\texttt{basic\_original}) and the corrected label (\texttt{basic\_corrected}), along with the confidence score and the reason for correction. Includes video paths and dialogue IDs for backtracking.
\end{itemize}

\subsection{Annotation Fields Hierarchy(Part of streaming data)}

\paragraph{Video-Level Interpretation}
Designed for context understanding:
\begin{itemize}
    \item \texttt{summary}: Abstract of the video content.
    \item \texttt{overall\_emotion\_trend}: Describes the dominant emotion and its evolution.
    \item \texttt{utterances}: A list containing multimodal evidence, intensity, and bilingual transcripts.
\end{itemize}

\paragraph{Sentence-Level Timestamps}
Designed for alignment:
\begin{itemize}
    \item Top-level metadata combined with a \texttt{segments} list (index, start/end time, text, slice filename).
\end{itemize}

\paragraph{Text-Based Multi-Annotation}
Designed for label reliability:
\begin{itemize}
    \item \textbf{Basic-7:} Aggregates individual annotations into \texttt{ds\_label} (consensus), \texttt{ds\_confidence} (reliability score), and \texttt{unique\_labels} (disagreement indicator).
    \item \textbf{GoEmotions:} Derived fields include label union/intersection, set size, and soft probabilities defined as $P(label) = \frac{\text{count}}{3}$.
    \item \textbf{Correction/Fusion:} Maps GoEmotions probabilities back to Basic-7, generating fields for the best candidate (\texttt{basic\_Goemotions\_best}), the final corrected label, and explicit flags for whether the original label was modified and why.
\end{itemize}

\subsection{Utilization Strategy}

To maximize the utility of the EmoS dataset, we recommend the following strategies based on the data structure:

\begin{itemize}
    \item \textbf{Single-Label Training (Basic-7):} Use \texttt{ds\_label} as the primary supervision signal. The \texttt{ds\_confidence} score can serve as a sample weight or a threshold for curriculum learning. The \texttt{unique\_labels} count identifies "hard samples" with high human disagreement.
    \item \textbf{Multi-Label Training (GoEmotions):} Utilize the soft probability vectors (derived from annotator agreement) rather than binary targets to model emotional ambiguity.
    \item \textbf{Multimodal Alignment:} For the \texttt{streaming} subset, use the timestamp JSON to align text transcripts with visual/audio slices. The utterance-level interpretations can serve as a test set for long-context understanding.
    \item \textbf{Data Cleaning:} Use \texttt{video\_id}/\texttt{clip\_id} (or \texttt{dia}/\texttt{utt} IDs) to join text-level aggregation results with video-level raw files.
    \item \textbf{Stratified Evaluation:} We recommend splitting train/validation sets based on \texttt{ds\_confidence} levels (High/Medium/Low) to ensure models are tested on samples with varying degrees of difficulty.
\end{itemize}

\section{Classification of Goemotions}
The classification of Goemotions is 28 categories, with an additional category of "neutral", which is consistent with the original paper.
Admiration, Amusement, Approval, Caring, Desire, Excitement, Gratitude, Joy, Love, Optimism, Pride, Relief, Anger, Annoyance, Disappointment, Disapproval, Disgust, Embarrassment, Fear, Grief, Nervousness, Remorse, Sadness, Confusion, Curiosity, Realization, Surprise, Neutral.

\section{Detailed Related Work}
\label{app:related_work}

In this section, we provide a detailed analysis of the evolution of MER datasets, categorizing them into laboratory-controlled, in-the-wild, and LLM-generated benchmarks.

\subsection{Lab-Controlled Datasets}
Early research predominantly relied on datasets collected in controlled laboratory environments \citep{mckeown2012semaine,ringeval2013recola}. A representative benchmark, \textbf{IEMOCAP} \citep{busso2008iemocap}, comprises approximately 12 hours of dyadic interaction. Due to its high-quality audio-visual recordings and fixed camera angles, it effectively captures rich facial expressions and has long served as a standard benchmark. Another significant dataset, \textbf{DAIC-WOZ} \citep{gratch2014daic}, focuses on clinical psychology scenarios through a ``virtual interviewer'' setup. However, these datasets universally suffer from a lack of ecological validity \citep{dhall2013emotiw}. Constrained by laboratory settings and scripted protocols, the emotional expressions lack spontaneity, limiting model generalization in unconstrained environments.

\subsection{In-the-wild Datasets: TV \& Social Media}
To pursue more naturalistic expressions, the community shifted towards datasets derived from TV shows and social media \citep{dhall2013emotiw,mollahosseini2017affectnet}. \textbf{MELD} \citep{poria2019meld}, extended from Friends, introduced multi-party interactions. However, it contains significant noise \citep{kanluan2023inwild_assessment}: audio tracks are contaminated by canned laughter, and cinematic editing results in frequent shot transitions. Such visual discontinuity bottlenecks multimodal feature extraction.

Similarly, \textbf{CMU-MOSI} \citep{zadeh2016mosi} and \textbf{CMU-MOSEI} \citep{zadeh2018mosei} provide large-scale monologues from YouTube. However, due to the lack of rigorous cleaning, MOSEI contains many non-emotional segments (e.g., neutral narration), and inconsistent label quality compromises its reliability. Addressing the scarcity of non-English resources, \textbf{CH-SIMS v2} \citep{liu2022chsims2} provided Chinese multimodal data with high-quality alignment but only offers coarse-grained (Positive/Negative) polarity labels, limiting its application in complex emotion analysis.

\subsection{LLM-Generated Datasets \& Challenges}
With the advancement of LLMs, works such as \textbf{Emotion-LLaMA} \citep{cheng2024emotionllama} and \textbf{MER-Caption} \citep{lian2025affectgpt} explore instruction tuning using synthetic data. These datasets introduce complex categories and natural language descriptions. However, they present two core pitfalls: (1) open-ended labels complicate standardized evaluation \citep{huang2023hallucination_survey}, and (2) excessive reliance on AI-generated annotations without rigorous human verification leads to inherent hallucination issues \citep{ji2023selfreflection}, compromising ground truth credibility.
\subsection{Other Dynamic Datasets}
Recent works like \textbf{DFEW} \citep{jiang2020dfew} and \textbf{MAFW} \citep{liu2022mafw} have scaled up dynamic facial expression recognition in the wild. However, these datasets primarily focus on visual modalities and lack the rigorous bilingual textual alignment and fine-grained psychological annotation (e.g., GoEmotions) present in EmoS.

\section{Prompt}
\label{sec:prompt}
\paragraph{Prompt A (Compact JSON Classification).}
\textbf{Instructions:}
\begin{itemize}
  \item Examine visual cues (facial expressions, gestures), vocal tone, and the transcript.
  \item Ignore background characters unless they directly drive the main speaker's emotion.
  \item Prefer ``Neutral'' only when no consistent emotion is detectable.
  \item If the transcript conflicts with the audio tone or visuals, prioritize audio/video evidence.
  \item In addition to a single primary emotion, estimate multi-label GoEmotions signals.
        Use \textbf{only} these labels: \{\texttt{Goemotions\_labels}\}.
  \item List every GoEmotions label that truly appears; include only plausible entries sorted by confidence.
  \item Return \textbf{JSON only} with the following fields:
\end{itemize}

\paragraph{Output Format (Text Only).}
Return \textbf{plain text} (not JSON) with the following fields:

\begin{itemize}
  \item \textbf{Predicted emotion:} \texttt{<label>}
  \item \textbf{Model confidence:} a float in \texttt{[0, 1]}
  \item \textbf{GoEmotions labels:} list all inferred labels (use only the allowed set), each with a confidence score in \texttt{[0, 1]}, sorted by confidence
  \item \textbf{Evidence:} one sentence describing the audio/visual/transcript cues used
  \item \textbf{GoEmotions rationale:} briefly explain why the multi-label set was chosen
\end{itemize}

\textbf{Summary reminder:}
\begin{itemize}
  \item Report only the GoEmotions labels you genuinely infer; there is no required minimum count.
  \item When transcripts disagree with tone or visuals, audio and video cues take priority.
\end{itemize}

\textbf{Clip metadata:}
\begin{itemize}
  \item language: \{\texttt{language}\}
  \item transcript: \{\texttt{transcript}\}
\end{itemize}

\paragraph{Prompt B (Streaming Utterance-Level Annotation).}
You are an emotion researcher specialized in multimodal analysis. Based on the uploaded video (including both visuals and audio), complete the following tasks:
\begin{enumerate}
  \item Summarize the clip in 2--3 sentences, highlighting key turning points and the overall emotional atmosphere.
  \item Analyze only the absolute main subject character in the visuals and storyline, ignoring all other characters or background figures. Extract all utterances spoken by the main subject (infer/supplement when necessary), and provide structured emotion annotations for each utterance.
    \begin{itemize}
      \item The primary emotion (\texttt{primary\_7class}) must be strictly one of the traditional seven classes:
      \texttt{Anger, Disgust, Fear, Joy, Neutral, Sadness, Surprise}.
      \item The secondary GoEmotions labels (\texttt{secondary\_Goemotions}) can be multi-label selections from the following list; if nothing matches, use \texttt{Unsure}:
      \texttt{admiration, amusement, approval, caring, desire, excitement, gratitude, joy, love, optimism, pride, relief, anger, annoyance, disappointment, disapproval, disgust, embarrassment, fear, grief, nervousness, remorse, sadness, surprise, confusion, curiosity, realization, neutral, Unsure}.
    \end{itemize}
  \item Output a UTF-8 JSON in the following format:
\end{enumerate}

\paragraph{Output Format (Text Only).}
\begin{enumerate}
  \item \textbf{Summary (2--3 sentences):} Briefly describe the clip, highlighting key turning points and the overall emotional atmosphere.
  \item \textbf{Overall Emotions Trend:}
  \begin{itemize}
    \item \textbf{Dominant emotion:} One label from the 7-class set (Anger, Disgust, Fear, Joy, Neutral, Sadness, Surprise).
    \item \textbf{Trajectory:} Describe how emotions evolve over time, citing MM:SS timestamps.
  \end{itemize}
  \item \textbf{Utterance-Level Annotations (chronological, non-overlapping timestamps):} For each utterance spoken by the main subject, provide:
  \begin{itemize}
    \item \textbf{Timestamp range:} MM:SS--MM:SS
    \item \textbf{Speaker guess (optional):} May be left empty
    \item \textbf{Chinese transcript:} The utterance in Chinese
    \item \textbf{English transcript (optional):} English transcription/translation if possible
    \item \textbf{Primary emotion (7-class):} Exactly one label from the 7-class set
    \item \textbf{Secondary emotions (GoEmotions):} A multi-label set chosen only from the allowed GoEmotions list; use \texttt{Unsure} if none fit
    \item \textbf{Intensity:} 1--5
    \item \textbf{Audio cues:} Tone, volume, pauses, etc.
    \item \textbf{Visual cues:} Facial/body cues, camera context, etc.
    \item \textbf{Evidence:} One brief sentence citing supporting cues from visuals and/or dialogue
  \end{itemize}
\end{enumerate}

\textbf{Constraint:} Output should be plain text (not JSON). If exact timestamps cannot be identified, explicitly state the assumptions used.

\textbf{Constraints:} Please output \textbf{UTF-8 JSON only}. Ensure each utterance's timestamp range does not overlap and is ordered chronologically. If exact timestamps cannot be identified, clearly state the assumptions used.

For Emotion Llama, we used the official sentiment classification prompt.
\end{document}